\documentclass[runningheads]{llncs}
\usepackage{cite}
\usepackage{graphicx}
\usepackage{cite}
\usepackage{latexsym}
\usepackage{hyperref}
\usepackage{amsmath,amssymb,amsfonts}
\usepackage{booktabs}
\usepackage{multirow}
\usepackage{floatrow}
\newfloatcommand{capbtabbox}{table}[][\FBwidth]
\usepackage{blindtext}
\begin{document}

\title{Cross-Task Representation Learning for Anatomical Landmark Detection
% Cross Task Knowledge Transfer for Anatomical Landmark Detection
%\thanks{Supported by organization x.}
}
\author{
Zeyu Fu\inst{1} \and
Jianbo Jiao\inst{1} \and
Michael Suttie \inst{2}\and
J. Alison Noble \inst{1}
}

\authorrunning{Z. Fu et al.}

\institute{
Department of Engineering Science, University of Oxford, Oxford, UK \\ \email{zeyu.fu@eng.ox.ac.uk}
\and
Nuffield Department of Women's and Reproductive Health, University of Oxford, Oxford, UK
}
\maketitle              % typeset the header of the contribution
\begin{abstract}
%Fetal alcohol syndrome (FAS) caused by prenatal alcohol exposure can result in a series of cranio-facial anomalies, and behavioural and neurocognitive problems. Current diagnosis of FAS is closely related to a set of facial characteristics, which are often obtained by manual examination.
%In many cases, diagnosis is heavily reliant on the recognition of a set of characteristic facial features, which can be subtle and difficult to objectively identify.
%To automatically detect the presence of FAS associated facial anomalies, we embark on anatomical landmark detection to facilitate the feature extraction.
Recently, there is an increasing demand for automatically detecting anatomical landmarks which provide rich structural information to facilitate subsequent medical image analysis.
%This motivates us to seek an automated facial screening tool to detect the presence of FAS associated facial anomalies, for which we embark on anatomical landmark detection as it provides rich geometric information for subsequent facial analysis.
%In this paper, we aim to extend the success of these well-trained models our facial shape analysis
%We address this task using CNN based heatmap regression, since it has achieved widely success in keypoint localization, due to its advantage of preserving high spatial resolution.
%Facial landmark detection has towards face perception facilitate various face analysis tasks
Current methods related to this task often leverage the power of deep neural networks, while a major challenge in fine tuning such models in medical applications arises from insufficient number of labeled samples.
%Current methods related to this task in computer vision often leverage the power of deep neural networks, while a major challenge in the deployment of such models in medical applications arises from insufficient number of labeled samples.
%To this end, we embark on facial landmark detection as it can provide rich geometric and anatomical information for various face analysis tasks.
%Current convolutional neural networks based heatmap regression methods has led to important advances in localizing facial landmarks, while a major challenge in the deployment of such models in medical applications arises from insufficient number of labeled samples.
%Recent development of convolutional neural works has led to important advances in keypoint localization, while a major bottleneck in the deployment of such models in medical applications arises from insufficient number of labelerd samples.
To address this, we propose to regularize the knowledge transfer across source and target tasks through cross-task representation learning. The proposed method is demonstrated for extracting facial anatomical landmarks which facilitate the diagnosis of fetal alcohol syndrome. The source and target tasks in this work are face recognition and landmark detection, respectively.
The main idea of the proposed method is to retain the feature representations of the source model on the target task data, and to leverage them as an additional source of supervisory signals for regularizing the target model learning, thereby improving its performance under limited training samples.
Concretely, we present two approaches for the proposed representation learning by constraining either final or intermediate model features on the target model.
Experimental results on a clinical
face image dataset demonstrate that the proposed approach works well with few labeled data, and outperforms other compared approaches.
%Experimental evaluations on a clinical face imaging dataset demonstrate the proposed approach can effectively transfer knowledge across related tasks, and outperform standard transfer learning approaches.
\end{abstract}
\section{Introduction}
\label{Introduction}
%Prenatal alcohol exposure (PAE) (For Mike).
%Formal diagnosis at the earliest possible stage is paramount, as it allows
%early intervention and reduces the risk of secondary disabilities.
%Fetal alcohol syndrome (FAS), as the most severe phenotype, is commonly associated with three classic facial features, which are short palpebral fissure length (PFL), a smooth philtrum and a thin upper lip \cite{FAS}.
%Conventional approaches towards extracting such anatomical measurements mostly rely on manual examination, which is tedious and subject to inter-operator variability.
%This necessitates the development of computer-aided facial assessment which would assist clinicians objectively making diagnoses.
%To design such an automated facial screen system,
Accurate localization of anatomical landmarks plays an important role for medical image analysis and applications such as image registration and shape analysis \cite{regression-voting}. It also
has the potential to facilitate the early diagnosis of Fetal Alcohol Syndrome (FAS) \cite{DSNT}. An FAS diagnosis requires the identification of at least 2 of 3 cardinal facial features; a thin upper lip, a smooth philtrum and a reduced palpebral fissure length (PFL)\cite{FASD}, which means that even a small inaccuracy in the PFL measurement can easily result in misdiagnosis.
%Given that 2 of 3 cardinal features are required for diagnosis means that even a small inaccuracy in the PFL measurement can easily result in misdiagnosis.
Conventional approaches for extracting anatomical landmarks mostly rely on manual examination, which is tedious and subject to inter-operator variability. To automate landmark detection, recent methods in computer vision \cite{stacked-hourglass,simple-baseline,mobileFAN} and medical image analysis \cite{attention-guided-landmark,regression-voting,DSNT} have extensively relied on convolutional neural networks (CNN) for keypoint regression.
Although these models have achieved promising performance, this task still remains challenging especially when handling the labeled data scarcity in medical domain, due to expensive and inefficient annotation process.
%because annotating such abundant landmarks is inefficient and requires extensive expert knowledge.
Transfer learning, in particular fine-tuning pre-trained models from similar domains have been widely used to help reduce over-fitting by providing a better initialization  \cite{transfer-learning}.
However, merely fine-tuning the existing parameters may arguably lead to a suboptimal local minimum for the target task, because much knowledge of the pre-trained model in the feature space is barely explored \cite{Inductive-TL,DELTA}.
To address this, we explore the following question: \emph{Is it possible to leverage the abundant knowledge from a domain-similar source task to guide or regularize the training of the target task with limited training samples?}

We investigate this hypothesis via cross-task representation learning, where ``cross-task" here means that the learning process is made between the source and target tasks with different objectives.
In this work, the proposed cross-task representation learning approach is illustrated for localizing anatomical landmarks in clinical face images to facilitate early recognition of fetal alcohol syndrome \cite{PFL}, where the source and target tasks are face recognition and landmark detection.
%Inspired by knowledge distillation, the proposed learning approach aims to retain the predictions of a domain-similar source model on the target task, and leverage them as additional source of supervisory signals for regularizing the target model learning, so as to improve its generalization ability under limited training samples.
Intuitively, the proposed representation learning is interpreted as preserving feature representations of a source classification model on the target task data, which serves as a regularization constraint for learning the landmark detector. Two approaches for the proposed
representation learning are developed by constraining either final or intermediate network features on the target model.
%Furthermore,  we consider a general scenario, where the source task data is not accessible.
%In particular, we investigate the possibility of regularizing the optimization of landmark detection by explicitly preserving the knowledg e from a rich  classification model with the ability.
%The proposed work allows to extend the success of strong models on a well-developed task to a domain-similar task with limited training samples, outperform the standard fine tuning. Preserving outputs on a related task is a more interpretable way to retain the important shared structures learned for the previous tasks.
%This capacity is useful when the new domain has limited training samples,and the source domain data is not accessible to student model.
%We evaluate the proposed method on a small and compare it with
\subsubsection{Related Work.}
Current state-of-art methods formulate the landmark detection as a CNN based regression problem, including two main frameworks: direct coordinate regression \cite{facial-attribute,wing-loss} and heatmap regression \cite{stacked-hourglass,simple-baseline}. Heatmap regression usually outperforms its counterpart as it preserves high spatial resolution during regression.
%and uses the predicted heatmaps to infer the landmark coordinates.
In medical imaging, several CNN architectures have been developed based on attention mechanisms \cite{regression-voting,attention-guided-landmark}, and cascaded processing \cite{Two-Stage-Task-Oriented} for the enhancement of anatomical landmark detection.
However, the proposed learning approach in this paper focuses on internally enriching the feature representations for the keypoint localization without complicating the network design.
%For medical imaging, Huang et al. \cite{DSNT} incorporated a differentiable spatial to numerical transform layer to a encoder-decoder network, which enables the model to directly output numerical coordinates without any postprocessing.
%Zhong et al. \cite{attention-guided-landmark} developed a two-stage U-Net with  attention guidance to localize anatomical
%landmarks in cephalometric X-ray images.
%Vlontzos et al. \cite{regression-DQN} introduced multi-agent reinforcement learning for multiple landmark detection in various medical imaging applications by designing a collaborative Deep Q-Network.
%However, current literature barely considered reusing a pre-trained model's knowledge to guide the learning process of anatomical landmark detection, which is studied in this paper.

Among existing knowledge transfer approaches, fine-tuning \cite{simple-baseline}, as a standard practice initializes from a pre-trained model and shifts its original capability towards a target task, where a small learning rate is often applied and some model parameters may need to be frozen to avoid overfitting.
However, empirically modifying the existing parameters may not generalize well over the small training dataset.
%In addition, multitask learning \cite{facial-attribute} integrates other related auxiliary tasks to facilitate the desired task in a joint learning fashion, requiring data and labels of all tasks are available during training, while such a requirement is hardly fulfilled in the clinical practice due to the concerns of privacy and storage.
%Our learning approach is in spirit similar to multitask learning, but we do not require access to the data from source domain.
%Each task with extra side information acts as a regularization step to learn a shared representation which is expected to benefit the primary task.
%However, solely slowing down the learning process may not explicitly exploit previously learned knowledge \cite{LWF}.
%which restrict acces  s to data in the source domain tasks access to data and labels from the source domain task.
%Progressive network \cite{PNN} Parameters for the original network remain unchanged, and newly added layers are connected to the existing one, whereas it unavoidably introduced extra number of parameters to be trained from scratch and rely on it at test time.
Knowledge distillation originally proposed for model compression \cite{knowledge-distillation} is also related to knowledge transfer.
%It is considered as a more interpretable way to retain the learning experiences from a well-trained model by designing teacher-student networks.
%The idea is to train a compact model reaching to the teacher performance  by encouraging its final predictions to be close to the teacher's predictions.
This technique has been successfully extended and applied to various applications, including hint learning \cite{hint-learning}, incremental learning \cite{LWF,LWM}, privileged learning \cite{privileged-info}, domain adaptation  \cite{cross-modality} and human expert knowledge distillation \cite{gaze-distillation}.
%Incremental learning with knowledge distillation continuously learns to solve new problems whilst maintaining its original capabilities,
%including image classification \cite{LWF,LWM}, sematic segmentation \cite{IL-Segmentation} and object detection \cite{incremental-detection},
%In addition, Patra et al.  proposed a new variant of privileged learning with knowledge distillation by leveraging human expert knowledge as addition input to enrich the teacher learning.
These distillation methods focused on training a compact model by operating the knowledge transfer across the same tasks \cite{knowledge-distillation,gaze-distillation,hint-learning}.
%require paired data from the source and target domain \cite{cross-modality}.
However, our proposed learning approach aims to regularize the transfer learning across different tasks.
\subsubsection{Contributions.}
%In this paper, we contribute a generalized version of teacher-student knowledge transfer for anatomical landmark detection under limited training data.
We propose a new deep learning framework for anatomical landmark detection under limited training samples.
The main contributions are:
(1) we propose a cross-task representation learning approach whereby the feature representations of a pre-trained classification model are leveraged for regularizing the optimization of landmark detection.
%To the best knowledge, our study is the first attempt to achieve the anatomical landmark detection by performing cross-task knowledge transfer.
(2) We present two approaches for the proposed representation learning by constraining either final or intermediate network features on the target task data. In addition, a cosine similarity inspired by metric learning is adopted as a regularization loss to transfer the relational knowledge between tasks.
(3) We experimentally show that the proposed learning approach performs well in anatomical landmark detection with limited training samples and is superior to standard transfer learning approaches.

\section{Method}
\label{Method}
In this section, we first present the problem formulation of anatomical landmark detection, and then describe the design of the proposed cross-task representation learning to address this task.
%Given an input image, our goal is to localize a set of pre-defined anatomical landmarks on the face.
%We firstly describe how a cross-task teacher model is obtained.
%\vspace{-0.2cm}
\subsection{Problem Formulation}
In this paper, our target task is anatomical landmark detection, which aims to localize a set of pre-defined anatomical landmarks given a facial image.
Let $\mathcal{D}^{t}= \{\mathbf{I}_{i}^{t}, \mathbf{p}_{i}^{t}\}_{i=1}^{N_{t}}$ be the training dataset with $N_{t}$ pairs of training samples in the target domain. $\mathbf{I}_{i}^{t}\in \mathbb{R}^{H \times W \times 3}$ represents a 2D RGB image with height $H$ and width $W$,
$\mathbf{p}_{i}^{t} = [(x_{1}, y_{1}), (x_{2}, y_{2}),...,(x_{K}, y_{K})] \in \mathbb{R}^{2\times K}$ denotes the corresponding labeled landmark coordinates, and $K$ is the number of anatomical landmarks ($K=14$).
%Note that the prediction target is no longer the classification labels as in the source task, thus the goal is to localize a set of pre-defined landmarks on each image.
We formulate this task using heatmap regression, inspired by its recent success in keypoint localization \cite{stacked-hourglass,simple-baseline}.
Following prior work \cite{stacked-hourglass}, we downscale the labeled coordinates to $1/4$ of the input size ($\mathbf{p}_{i}^{t}= \mathbf{p}_{i}^{t}/4$), and then transform them to a set of heatmaps $\mathbf{G}_{i}^{t} \in \mathbb{R}^{(H/4) \times (W/4) \times K}$. Each heatmap $\mathbf{g}_{k}^{t} \in \mathbb{R}^{(H/4) \times (W/4)}, k\in \{1,...,K\}$ is defined as a 2D Gaussian kernel centered on the $k$-th landmark coordinate $(x_{k},y_{k})$. The $(a,b)$ entry of $\mathbf{g}_{k}^{t}$ is computed as $\mathbf{g}_{k}^{t}(a,b)=\exp (-\frac{(a-x_{k})^{2}+(b-y_{k})^{2}}{2 \sigma^{2}})$, where $\sigma$ denotes the kernel width ($\sigma = 1.5$ pixels).
Consequently, the goal is to learn a network which regresses each input image to a set of heatmaps, based on the updated dataset $\mathcal{D}^{t}= \{\mathbf{I}_{i}^{t}, \mathbf{G}_{i}^{t}\}_{i=1}^{N_{t}}$.

For this regression problem, most state-the-of-the-art methods \cite{simple-baseline,mobileFAN} follow the encoder-decoder design, in which a pre-trained network (e.g. ResNet50 \cite{resnet}) is usually utilized in the encoder for feature extraction, and then the entire network or only the decoder is fine-tuned during training. However, due to the limited number of training samples in our case, merely relying on standard fine-tuning may not always provide a good localization accuracy. Therefore, we present the proposed solution to address this problem in the next section.
%or require exhaustive search for a set of ad-hoc hyper-parameters.
\subsection{Cross-Task Representation Learning}

\subsubsection{Overview.} 
\begin{figure}[t]
\centering
\setlength{\abovecaptionskip}{-3pt}
\setlength{\belowcaptionskip}{-5pt}
\includegraphics[width=1\linewidth]{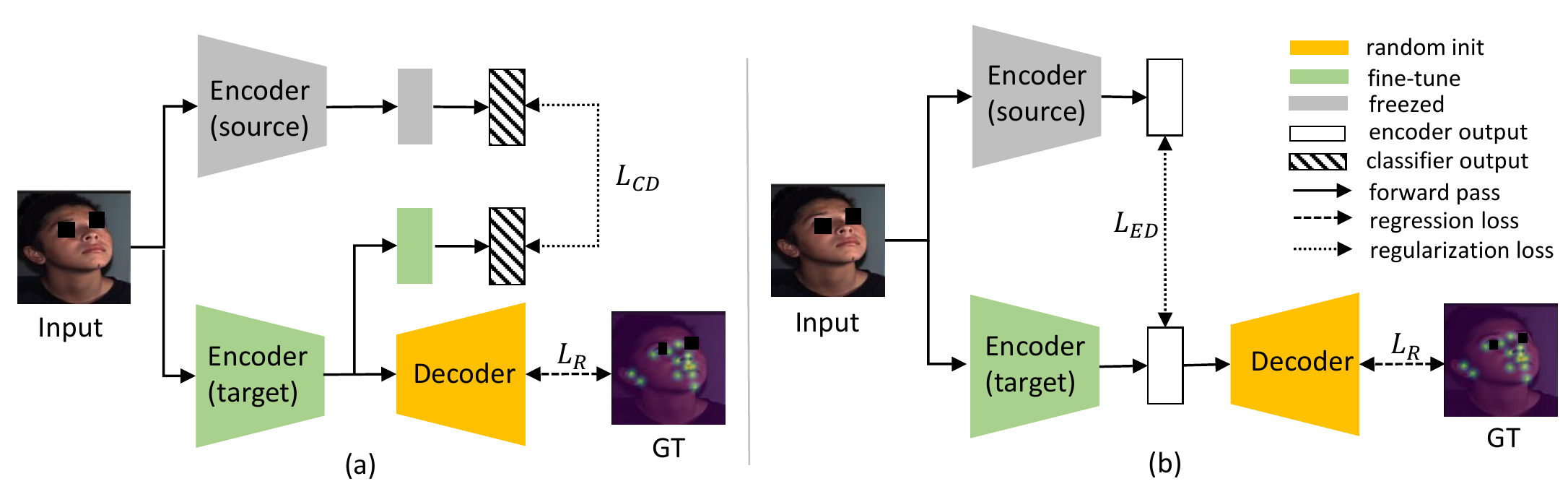}
\caption{Illustration of proposed approaches for learning the anatomical landmark detection models, where (a) presents the regularization constraint on the final layer output ($L_{CD}$), and (b) is to constrain the predictions on the encoder output ($L_{ED}$).}
\label{overall-system}
\end{figure}

Fig. \ref{overall-system} depicts the overall design of the proposed cross-task representation learning approach.
Firstly, the source model pre-trained on a face classification task is operated in the inference mode to predict rich feature representations from either classification or intermediate layers for the target task data. The target model is then initialized from the source model and extended with a task-specific decoder for the task of landmark detection ($L_{R}$).
Obtained feature representations are then transferred by regularization losses ($L_{CD}$ or $L_{ED}$) for regularizing the target model learning.
%Main steps of the proposed framework are detailed in the following subsections.
%The main idea is to incorporate teacher's predictions as additional interpretations about current training samples to regularize the target model designated for heatmap regression.
%This is because such well-trained networks provide rich shared representations about facial attributes, which are useful to promote other related tasks \cite{self-supervised}.
%\vspace{-0.3cm}
\subsubsection{Source Model.}
%Prior works in knowledge distillation require the teacher and student networks both to have same prediction targets, so that the student for the desired task can which restricts its potential to which relaxes this restriction, and which fills in the gap of distilling knowledge across related tasks. The teacher network may need to be repurposed for the target task, which
%Prior works in knowledge distillation require that the teacher and student networks have the same learning targets. Here, we aim to relax this requirement
%which means in this case the teacher network has to adapt  so that the distilled knowledge can be transferred between both ends guide the student learning
We consider a pre-trained face classification network as our source model,
since generic facial representations generated from this domain-similar task have been demonstrated to be helpful for other facial analysis \cite{self-supervised}.
%Here, we aim to extract the predictions from the teacher network
%Considering a face classification task with $C$ classes,
Formally, let $\mathcal{S}_{\theta_{1},\theta_{2}}: \mathbb{R}^{H \times W \times 3}\rightarrow \mathbb{R}^{C}$ be the source network for a face classification task with $C$ classes, where $\theta_{1}$ and $\theta_{2}$ are the learnable parameters. The network consists of a feature extractor (encoder) $f_{\theta_{1}}^{s}: \mathbb{R}^{H \times W \times 3}\rightarrow \mathbb{R}^{d}$ and a classifier $g_{\theta_{2}}^{s}: \mathbb{R}^{d} \rightarrow \mathbb{R}^{C} $, where $d$ denotes the dimensionality of the encoder output.
%Note that the prediction target is different from as in the source task, thus the goal is to localize a set of pre-defined landmarks on each image.
%let $\mathcal{D}^{(s)}= \{(\mathbf{I}_{i}^{(s)}, \mathbf{y}_{i}^{(s)})\}_{i=1}^{N_{s}}$ be the training dataset, where $N_{s}$ is the number of training samples in the source domain.
A cross-entropy loss is typically used to train the network $\mathcal{S}_{\theta_{1},\theta_{2}} := g_{\theta_{2}}^{s}(f_{\theta_{1}}^{s}(\mathbf{I}))$ which maps a facial image to classification scores based on a rich labeled dataset $\mathcal{D}^{s}$.
%and $\mathbf{y}_{i}^{(s)} \in \mathbb{R}^{C}$ is the corresponding one-hot label, where $C$ is the number of classes.
In practice, we adopt a pre-trained ResNet-50 \cite{resnet} model from VGGFace2 \cite{VGGFACE2} for the source network. Other available deep network architectures could also be utilized for this purpose.
%One interpretation is that these predictions as additional interpretations about the training samples are used to regularize the target task
%\vspace{-0.35cm}
\subsubsection{Target Model.}
For the task of heatmap regression, the target network $\mathcal{T}_{\theta_{1},\theta_{2}}$ is firstly initialized from the pre-trained source network.
We then follow the design of \cite{simple-baseline}, employing three deconvolutional layers after the encoder output $f_{\theta_{1}}^{t}(\mathbf{I})$ to recover the desired spatial resolution, where each layer has the dimension of 256 and $4 \times4$ kernel with the stride of $2$. Finally, a $1 \times 1$ convolutional layer is added to complete this task-specific decoder $h_{\theta_{3}}^{t}(f_{\theta_{1}}^{t}(\mathbf{I})): \mathbb{R}^{d}\rightarrow \mathbb{R}^{(H/4) \times (W/4) \times K}$.
%For a standard fine-tuning implementation, the classifier $g_{\theta_{2}}^{s}$ is no longer needed, so the student network can be reformulated as: $\mathcal{S}_{\theta_{1},\theta_{3}}=: h_{\theta_{3}}^{s}(f_{\theta_{1}}^{s}(\mathbf{I})) $.   The target is to train the decoder with the student encoder being either freezed or not. The loss function to train the model is given as,
The primary learning objective is to minimize the following loss between the decoder outputs and the labeled heatmaps,
\begin{equation}\label{regression-loss}
L_{R}= \frac{1}{N_{t}}\sum_{i=1}^{N_{t}} \left\|\mathbf{G}_{i}^{t}-h_{\theta_{3}}^{t}(f_{\theta_{1}}^{t}(\mathbf{I}_{i}^{t}))\right\|^{2}_{F}
\end{equation}
where $F$ denotes the Frobenius norm.
%\begin{equation}\label{regression-loss}
%L_{R}(\mathbf{G}_{i}^{s},\mathbf{I}_{i}^{s})= \left\|\mathbf{G}_{i}^{s}-h_{\theta_{3}}^{s}(f_{\theta_{1}}^{s}(\mathbf{I}_{i}^{s}))\right\|^{2}_{2}
%\end{equation}
\subsubsection{Regularized Knowledge Transfer.}
\begin{figure}[t]
\centering
\setlength{\abovecaptionskip}{-2pt}
\setlength{\belowcaptionskip}{-5pt}
\includegraphics[width=0.8\linewidth]{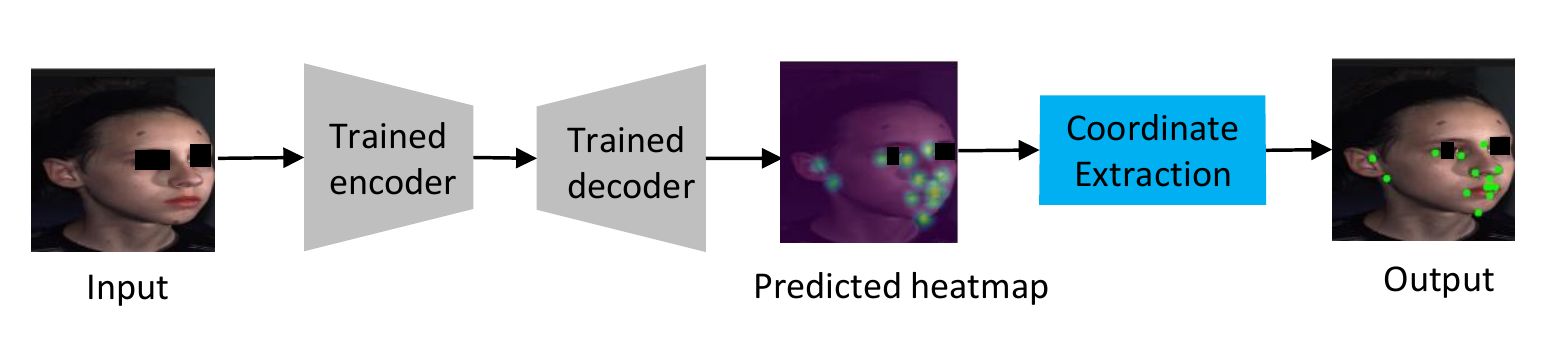}
\caption{Illustration of proposed framework for testing landmark detection models.}
\label{plot_inference}
\end{figure}
%To regularize the network training, standard knowledge distillation requires the cross-task teacher network to be re-trained on the target task data, so that the same prediction targets between networks can be aligned \cite{mobileFAN}. In contrast, the proposed approach relaxes this requirement and directly runs the pre-trained teacher network in the inference mode to generate predictions for the target task data $\mathcal{D}^{t}$, which are further transferred through a distillation loss $L_{KD}$ to regularize the current learning.
Motivated by knowledge distillation, we aim to regularize the network training by directly acquiring the source model's predictions for the target task data $\mathcal{D}^{t}$, which are further transferred through a regularization loss $L_{D}$.
Hence, the total loss is defined as,
\begin{equation}\label{total-loss}
L=L_{R}+ \lambda L_{D}
\end{equation}
where $\lambda $ is a weighting parameter. If $\lambda = 0$, the knowledge transfer becomes standard fine-tuning, as no regularization is included.

For the design of $L_{D}$, we firstly consider constraining the distance between the final layer outputs of the two networks, as shown in Fig. \ref{overall-system} (a). Similar to the distillation loss in \cite{knowledge-distillation}, we use a temperature parameter $\mu$ with $softmax$ function to smooth the predictions,  but the original cross-entropy function is replaced by the following term,
%For simplicity, we denote  concisely as and ignore the obtained detections on the detector index
%Prior works in knowledge distillation require the teacher and student networks both to have same prediction targets, so that the student can which restricts its potential to which relaxes this restriction, and which fills in the gap of distilling knowledge across related tasks. The teacher network may need to be repurposed for the target task, which
%\subsubsection{Distillation on the Classification Output}
%\subsubsection{Distillation on the Encoder Output}
%\subsection{Incremental Heatmap Regression}
%\label{Incremental Heatmap Regression}
%This is helpful for our subjects with partizcular age group and ethnic background to mitigate the dataset bias during knowledge transfer.
\begin{equation}\label{output-loss}
L_{CD}= \frac{1}{N_{t}}\sum_{i=1}^{N_{t}}\left\|softmax\left( \frac{g_{\theta_{2}}^{s}(f_{\theta_{1}}^{s}(\mathbf{I}_{i}^{t}))}{\mu}\right)-softmax\left( \frac{g_{\theta_{2}}^{t}(f_{\theta_{1}}^{t}(\mathbf{I}_{i}^{t}))}{\mu}\right)\right\|^{2}_{2}.
\end{equation}
The purpose of this design of $L_{CD}$ is to directly align the facial embeddings between instances, instead of preserving the original classification ability.
%\begin{equation}\label{encoder-loss}
%L_{CD}(\mathbf{I}_{i}^{s})= \left\|softmax\left( \frac{g_{\theta_{2}}^{t}(f_{\theta_{1}}^{t}(\mathbf{I}_{i}^{s}))}{\mu}\right)-softmax\left( \frac{g_{\theta_{2}}^{s}(f_{\theta_{1}}^{s}(\mathbf{I}_{i}^{s}))}{\mu}\right)\right\|^{2}_{2}
%\end{equation}

Moreover, we consider matching the features maps produced from both encoders as another choice, as shown in Fig. \ref{overall-system} (b). Motivated by the work in \cite{RKD}, we adopt the cosine similarity for the feature alignment as described below,
\begin{equation}\label{encoder-loss}
L_{ED}= 1-  \sum_{i=1}^{N_{t}}\cos (f_{\theta_{1}}^{s}(\mathbf{I}_{i}^{t}), f_{\theta_{1}}^{t}(\mathbf{I}_{i}^{t})).
\end{equation}
%\begin{equation}\label{output-loss}
%L_{ED}(\mathbf{I}_{i}^{s})= 1-  \sum_{i=1}^{N_{s}}\cos (f_{\theta_{1}}^{t}(\mathbf{I}_{i}^{s}), f_{\theta_{1}}^{s}(\mathbf{I}_{i}^{s}))
%\end{equation}
We conjecture that penalizing higher-order angular differences in this context would help transfer the relational information across different tasks, and also give more flexibility for the target model learning. Besides, both regularization terms can be combined together to regularize the learning process. Different approaches of the proposed learning strategy will be evaluated in the experimental section.
%\begin{equation}\label{distillation-loss}
%L_{KD}=\alpha L_{CD}+\beta L_{ED}
%\end{equation}
%\begin{equation}\label{customized-loss1}
%L_{ED} = -\sum_{x \in \mathbf{X}}y_{k}(x)\log(p_{k}(x))
%\end{equation}
%\begin{equation}\label{customized-loss2}
%L_{CD} = 1-\frac{2\sum_{x\in \mathbf{X}}y_{k}(x)p_{k}(x)}{\sum_{x\in \mathbf{X}}p^2_k(x)+\sum_{x\in \mathbf{X}}y^2_k(x)}
%\end{equation}
%
%\begin{equation}\label{customized-loss2}
%L_{CD} = 1-\frac{2\sum_{x\in \mathbf{X}}y_{k}(x)p_{k}(x)}{\sum_{x\in \mathbf{X}}p^2_k(x)+\sum_{x\in \mathbf{X}}y^2_k(x)}
%\end{equation}

During inference, as shown in Fig. \ref{plot_inference}, only the trained target model is used to infer the heatmaps, and each of them is further processed via an $argmax$ function to obtain final landmark locations.
%These representations, transferred by the proposed cross-task knowledge distillation are only utilized in the student learning process to explicitly provide an additional supervisory signal for regularizing the model parameters
%\subsubsection{Overview.}Firstly, the teacher model pre-trained on a face classification task is operated in the inference mode to generate predictions from either classification or intermediate layers for the target task data. The student model is then initialized from a clone of teacher model and extended with a task-specific decoder for the task of landmark detection.
%Teacher's predictions, transferred by distillation losses ($L_{CD}$ or $L_{ED}$)
%provide additional supervisory signals for regularizing the student learning.
%
%Main steps of the proposed framework are detailed in the following subsections.
\section{Experiments}
\label{Experiments}
\subsection{Dataset and Implementation Details}
We evaluate the proposed approach for extracting facial anatomical landmarks. Images used for training and test datasets were collected by the Collaborative Initiative on Fetal Alcohol Spectrum Disorders (CIFASD)\footnote{https://cifasd.org/}, a global multi-disciplinary consortium focused on furthering the understanding of FASD.
It contains subjects from 4 sites across the USA, aged between 4 and 18 years. Each subject was imaged using a commercially available static 3D photogrammetry system from 3dMD\footnote{http://www.3dmd.com/}. For this study, we utilize the high-resolution 2D images captured during 3D acquisition, which are used as UV mapped textures for the 3D surfaces.
%We evaluate the proposed approach for extracting facial anatomical landmarks. Images used for training and test datasets were collected by the Collaborative Initiative on Fetal Alcohol Spectrum Disorders (CIFASD), a global multi-disciplinary consortium focused on furthering the understanding of FASD.
%This dataset contains subjects from 4 sites across the USA, aged between 4 and 18 years. Each subject was imaged using a commercially available static 3D photogrammetry system from 3dMD \footnote{http://www.3dmd.com/}. For this study, we utilize the high-resolution 2D images captured during 3D acquisition, which are used as UV mapped textures for the 3D surfaces.

Specifically, we acquired in total 1549 facial images annotated by an expert, and randomly split them into training/validation set (80\%), and test set (20\%). All the images were cropped and resized to $256 \times 256$ for the network training and evaluation. Standard data augmentation was performed with randomly horizontal flip (50\%) and scaling ($0.8$).
During training, the Adam optimizer \cite{Adam} was used for the optimization with the mini-batch size of $2$ for $150$ epochs. A polynomial decay learning rate was used with the initial value of $0.001$. Parameters of $\lambda$ and $\mu$ used in (\ref{total-loss}) and (\ref{output-loss}) were set to $0.002$ and $2$, respectively.

\subsection{Evaluation Metrics}
\begin{figure}[t]
\centering
\setlength{\abovecaptionskip}{-1pt}
\setlength{\belowcaptionskip}{-5pt}
\includegraphics[width=0.9\linewidth]{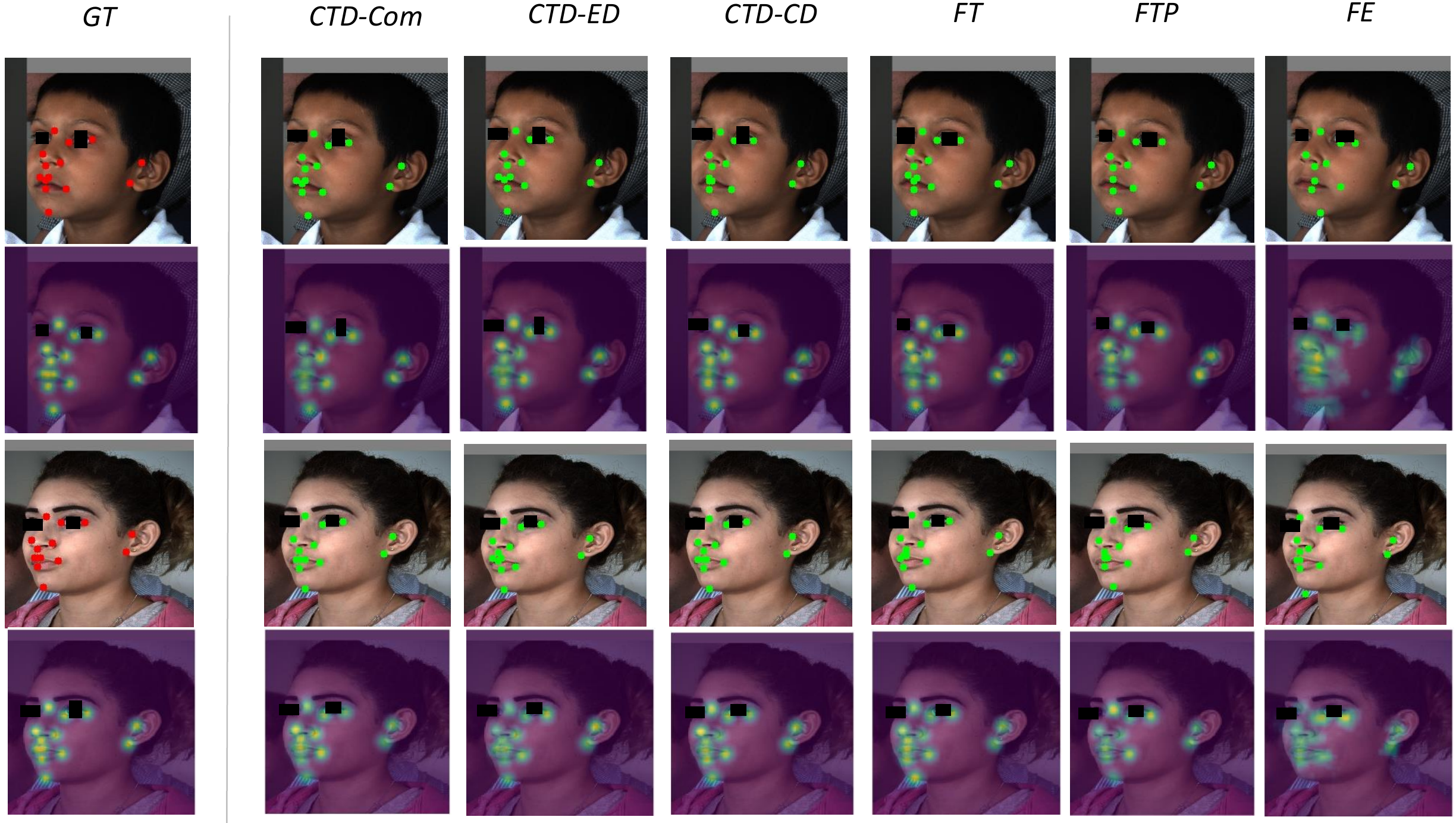}
\caption{Qualitative performance of landmark prediction and heatmap regression on the test set. Subjects' eyes are masked for privacy preservation. Better viewed in color.}
\label{qualitative_results}
\end{figure}

For the evaluation, we firstly employ the Mean Error (ME), which is a commonly-used evaluation metric in the task of facial landmark detection. It is defined as, $\mathrm{ME}=\frac{1}{N_{e}} \sum_{i=1}^{N_{e}} \frac{1}{K}\left\|\mathbf{p}_{i}-\hat{\mathbf{p}}_{i}\right\|_{2}$, where $N_{e}$ is the number of images in the test set, and $\mathbf{p}_{i}$ and $\hat{\mathbf{p}}_{i}$ denote the manual annotations and predictions, respectively.
Note that the original normalization factor measured by inter-ocular distance (Euclidean distance between outer eye corners) is not included in this evaluation, due to the unavailable annotations for the other eye, as illustrated in Fig. \ref{qualitative_results}.
In addition, we use the Cumulative Errors Distribution (CED)
curve with the metrics of Area-Under-the-Curve (AUC) and Failure Rate (FR), where a failure case is considered if the point-to-point Euclidean error is greater than $1.2$. Higher scores of AUC or lower scores of FR demonstrate the larger proportion of the test set is well predicted.
%All the experimental results were achieved by means of a work station with an Intel i7 CPU, a GeForce RTX 2080Ti GPU, and 64GB of RAM. Models compared in this study were trained with the same parameter settings for evaluation.

\begin{figure}[t]
\begin{floatrow}
\ffigbox{
\centering
\includegraphics[clip, trim=0.45cm 0.01cm 3.8cm 3.2cm, width=0.85\linewidth]{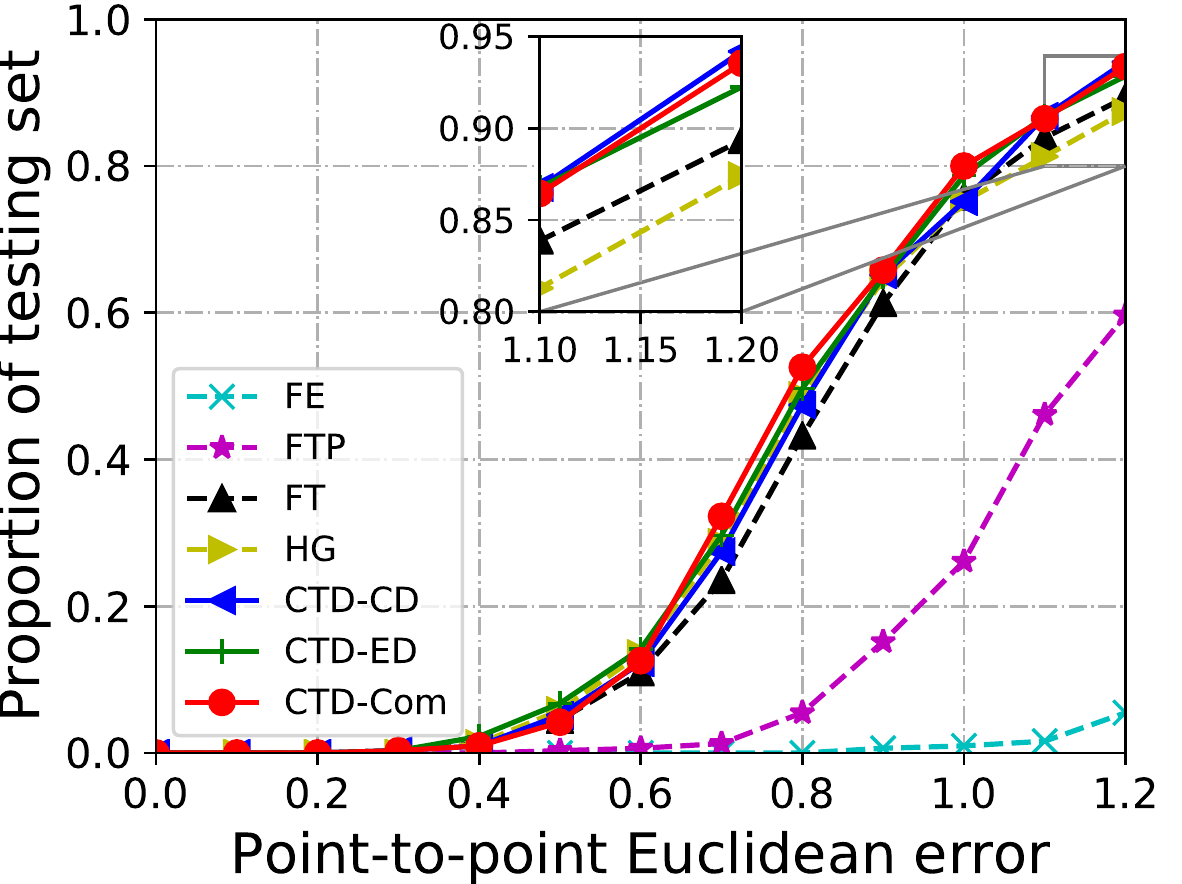}
}{\caption{Evaluation of CED curve on the test set. Better viewed in color.}
\label{ced-results}
}
\capbtabbox{
%\begin{table}[t]
\centering
\begin{tabular}{c|c|c|c}
\toprule
\multirow{2}{*}{Method}  & \multirow{2}{*}{ME $\pm$ SD}   & \multirow{2}{*}{FR} & \multirow{2}{*}{AUC}  \\
&&  &   \\

\hline
FE \cite{simple-baseline} & 1.822$\pm$0.501&  94.52\% & 0.01  \\
FTP \cite{simple-baseline} & 1.161$\pm$0.261&  40.32\%  & 0.10  \\
FT \cite{simple-baseline} & 0.858$\pm$\textbf{0.238}& 10.65\% &   0.29   \\
HG \cite{stacked-hourglass}& 0.879$\pm$0.386& 12.58\% &   0.30   \\
\hline
CTD-CD & 0.842$\pm$0.246 & \textbf{5.81\%} &  0.31  \\
CTD-ED & 0.830$\pm$0.245 & 7.74\%&  \textbf{0.32}  \\
CTD-Com & \textbf{0.829}$\pm$0.253 &6.45\% &  \textbf{0.32}  \\
\bottomrule
\end{tabular}
}{\caption{Quantitative evaluation on the test set. }
\label{quantative-results}}
%\end{table}
\end{floatrow}
\end{figure}
\subsection{Results and Discussions}
To verify the effectiveness of the proposed cross-task representation learning (CTD) approach, we compare to a widely-used CNN model: stacked Hourglass (HG) \cite{stacked-hourglass} and three variants of fine-tuning \cite{simple-baseline} without regularization ($\lambda = 0$): Feature Extraction (FE) with freezing the encoder, Fine Tuning Parts (FTP) without freezing the final convolutional layer of the encoder, and Fine Tuning (FT) without freezing any layer.
In addition, we present an ablation study to examine the significance of each approach in our proposed CTD, including the regularization on the classifier output (CTD-CD), the regularization on the encoder output (CTD-ED), and the regularization on both outputs (CTD-Com).
%For a fair comparison, methods presented in here were learned and evaluated using the same parameter configuration.
%The number of model parameters for inference is same for different proposed variants.

Fig. \ref{qualitative_results} shows the qualitative comparisons between different models on the test set. As we can see, the predicted landmarks from the proposed methods generally achieve the better alignment with the ground truth (the first left column) than the others, and seem to be more robust to difficult pixels especially when landmarks are in close proximity (upper lip).
One possible reason is that feature representations generated from the source model encode richer facial semantics, which make landmark spatial locations more discriminative.
Furthermore, the visualization of predicted heatmaps explains how each compared model responds to the desired task. We observe that our cross-task representation learning can effectively suppress spurious responses and improve the feature confidence in related regions, so that more accurate predictions can be achieved.
%In face classification, the key in distinguishing different identities is the appearance around the key-points such as shape and color, but the difference of key-point locations for different identities is tiny. As a result, face identity is not the main influencing factor for these locations, but it is still related as different identities may have slightly different locations.

On the other hand, Table \ref{quantative-results} summarizes the quantitative evaluation by reporting the statistics for each model. Fig. \ref{ced-results} depicts the CED curve which provides an intuitive understanding of the overall performance of the compared models.
These evaluations above demonstrate that the proposed methods consistently outperform standard fine-tuning solutions.
Moreover, CTD-ED performs slightly better than CTD-CD considering the scores of ME and AUC. This may be explained by the fact that features from intermediate layers are not only semantic, and also contain to some extent structural information which is beneficial for localization \cite{cross-modality}.
Interestingly, CTD-Com using both regularization losses achieves similar results in CTD-ED, as a result, CTD-ED may be considered as a better choice for the regularization of transfer learning.
\section{Conclusions}
\label{Conclusions}
%\begin{table}[t]
%\centering
%\setlength{\abovecaptionskip}{-1pt}
%%\setlength{\belowcaptionskip}{-0.1pt}
%\caption{Quantitative results of, \textbf{Bold} indicates the best results.}
%\begin{tabular}{|p{2cm}|p{2.5cm}|p{2.5cm}|p{1.5cm}|}
%\hline
%Method & ME$\pm$Std ($\downarrow$)  & Failure Rate ($\downarrow$) & AUC ($\uparrow$) \\
%\hline
%\hline
%FE & 1.822$\pm$0.501&  94.52\% & 0.01  \\
%FTP & 1.161$\pm$0.261&  40.32\%  & 0.10  \\
%FT & 0.858$\pm$0.238& 10.65\% &   0.29   \\
%CTD-CD & 0.842$\pm$0.246 & 5.81\% &  0.31  \\
%CTD-ED & 0.830$\pm$0.245 & 7.74\%&  0.32   \\
%CTD-Com & 0.829$\pm$253 &6.45\% &  0.32  \\
%\hline
%\end{tabular}
%\label{quantative-results}
%\end{table}
In this paper, we presented a new cross-task representation learning approach to address the problem of anatomical landmark detection where labeled training data is limited.
The proposed learning approach considered reusing the knowledge from a domain-similar source task as a regularization constraint for learning the target landmark detector.
Moreover, several regularization constraints for the proposed learning approach were considered.
Experimental results suggested that the proposed learning approach works well with limited training samples and outperforms other compared solutions.
The proposed approach can be potentially applied to other
related applications in the clinical domain where the target task has small training set and the source task
data is not accessible.
\subsubsection{Acknowledgements.}
This work was done in conjunction with the Collaborative Initiative on Fetal Alcohol Spectrum Disorders (CIFASD), which is funded by grants from the National Institute on Alcohol Abuse and Alcoholism (NIAAA).
%Additional information about CIFASD can be found at www.cifasd.org.
This work was supported by NIH grant U01AA014809 and EPSRC grant EP/M013774/1.

\bibliographystyle{splncs04}
\bibliography{mybib}

\end{document}